\documentclass[10pt,twocolumn,letterpaper]{article}

\usepackage{cvpr}              
\newcommand{\myparagraph}[1]{{\bf #1}}
\def\Ours{\textsc{CapsFusion}\xspace}
\def\dset{\textsc{CapsFus}-120M\xspace}
\def\model{\textsc{CapsFus}-LLaMA\xspace}

\usepackage{multirow}
\usepackage{pifont}
\usepackage{amsmath}
\usepackage{makecell}
\usepackage[accsupp]{axessibility}

\usepackage[dvipsnames]{xcolor}

\definecolor{cvprblue}{rgb}{0.21,0.49,0.74}
\definecolor{capsblue}{RGB}{231,239,252}
\definecolor{capsyellow}{RGB}{253,243,208}
\definecolor{capsred}{RGB}{251,214,211}
\newcommand{\authorskip}{\hspace{3mm}}
\usepackage[pagebackref,breaklinks,colorlinks,citecolor=cvprblue]{hyperref}

\def\paperID{10084} 
\def\confName{CVPR}
\def\confYear{2024}

\title{\Ours: Rethinking Image-Text Data at Scale\vspace{-3mm}}

\author{Qiying Yu\textsuperscript{1,2}\thanks{Equal contribution. $\dag$ Correspondence to \textit{wangxinlong@baai.ac.cn}. 
} \authorskip Quan Sun\textsuperscript{2}$^*$ \authorskip Xiaosong Zhang\textsuperscript{2} \authorskip Yufeng Cui\textsuperscript{2} \authorskip Fan Zhang\textsuperscript{2}  \authorskip \\
Yue Cao\textsuperscript{3} \authorskip Xinlong Wang\textsuperscript{2}$^\dag$ \authorskip Jingjing Liu\textsuperscript{1} \\[2mm]
{
\fontsize{10.4pt}{9.84pt}\selectfont
\textsuperscript{1} Institute for AI Industry Research (AIR), Tsinghua University \hspace{5mm} \textsuperscript{2} Beijing Academy of Artificial Intelligence}\\
{
\fontsize{10.4pt}{9.84pt}\selectfont
\textsuperscript{3} Independent Researcher}\\[2.5mm]
{
\fontsize{9.4pt}{9.84pt}\selectfont 
Code \& Dataset: \url{https://github.com/baaivision/CapsFusion}
}
\vspace{-1em}
}

\begin{document}
\maketitle
\begin{abstract}
Large multimodal models demonstrate remarkable generalist ability to perform diverse multimodal tasks in a zero-shot manner. Large-scale web-based image-text pairs contribute fundamentally to this success, but suffer from excessive noise. Recent studies use alternative captions synthesized by captioning models and have achieved notable benchmark performance. However, our experiments reveal significant Scalability Deficiency and World Knowledge Loss issues in models trained with synthetic captions, which have been largely obscured by their initial benchmark success. Upon closer examination, we identify the root cause as the overly-simplified language structure and lack of knowledge details in existing synthetic captions. To provide higher-quality and more scalable multimodal pretraining data, we propose \Ours, an advanced framework that leverages large language models to consolidate and refine information from both web-based image-text pairs and synthetic captions. Extensive experiments show that \Ours captions exhibit remarkable all-round superiority over existing captions in terms of model performance (\eg, 18.8 and 18.3 improvements in CIDEr score on COCO and NoCaps), sample efficiency (requiring 11-16 times less computation than baselines), world knowledge depth, and scalability. These effectiveness, efficiency and scalability advantages position \Ours as a promising candidate for future scaling of LMM training.
\end{abstract}

\definecolor{evaunit01green}{RGB}{82,208,83}
\newcommand{\evagreen}[1]{\textcolor{evaunit01green}{#1}}
\newcommand{\dtplus}[1]{\fontsize{6pt}{0.1em}\selectfont (\textbf{\evagreen{#1}})}

\section{Introduction}
\label{sec:intro}

Large Multimodal Models~\cite{alayrac2022flamingo,sun2023emu,liu2023llava} (LMMs), which as versatile multimodal generalists bridge powerful pretrained large language models~\cite{touvron2023llama2,touvron2023llama} and vision encoders~\cite{radford2021clip,sun2023evaclip}, have garnered significant success in zero-shot multimodal tasks.
Although image-text pairs harvested directly from the web~\cite{schuhmann2022laion5b} contribute instrumentally to the success of current LMMs, such web-scale data tend to be noisy and sub-optimal for model training~\cite{li2022blip,jia2021align}. Thus, strategies have been devised to harness synthetic captions generated by image captioning model~\cite{li2022blip}, which has augmented model performance notably~\cite{dong2023dreamllm,li2023blip2,sun2023emu,bai2023qwen,instructblip,ge2023seed-llama} by adopting large-scale synthetic caption datasets such as LAION-COCO~\cite{laioncoco} and BLIP-LAION~\cite{li2022blip}.

\begin{figure}
  \centering
    \includegraphics[width=\linewidth]{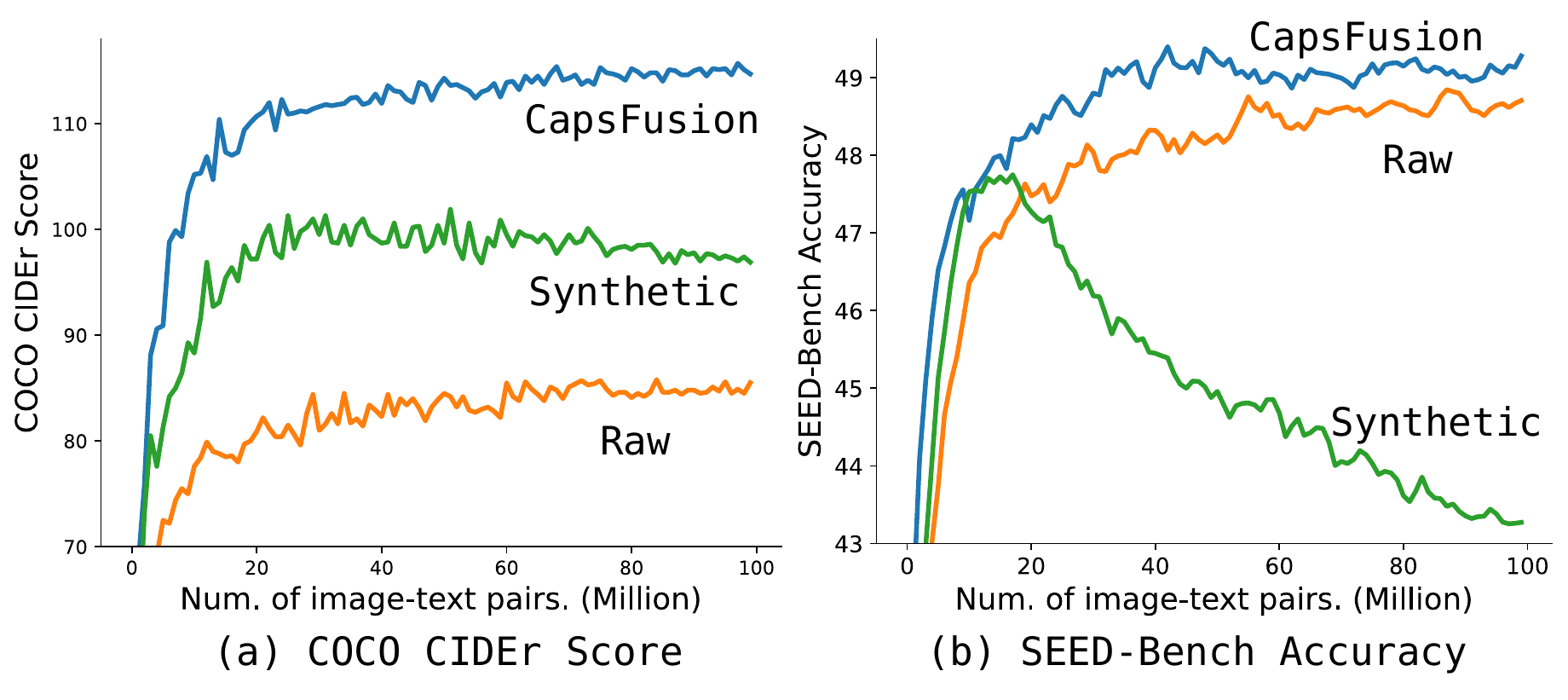}
  \vspace{-2em}
  \caption{Training process of models trained on different captions.
  }
  \label{fig:intro_scale_def}
\end{figure}
\begin{figure}
  \centering
  \vspace{-1em}
    \includegraphics[width=\linewidth]{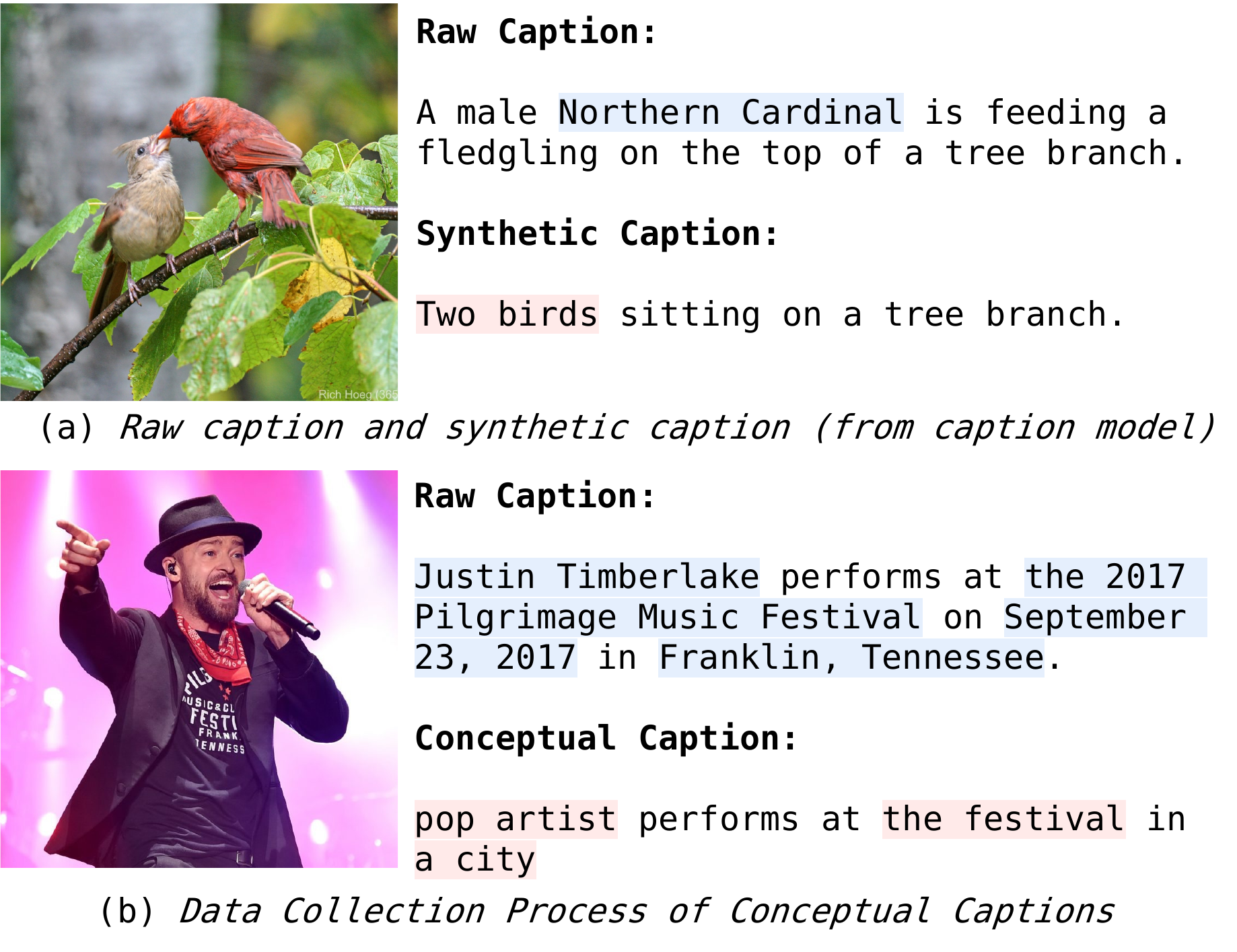}
\vspace{-2em}
  \caption{(a) Comparison of raw and synthetic captions for training. (b) Data processing of Conceptual Captions~\cite{sharma2018cc3m}, where \colorbox{capsblue}{real-world information} is substituted with \colorbox{capsred}{generic concepts}.
}
\vspace{-1em}
  \label{fig:intro_knowledge}
\end{figure}

\begin{figure*}[t]
  \centering
    \includegraphics[width=\linewidth]{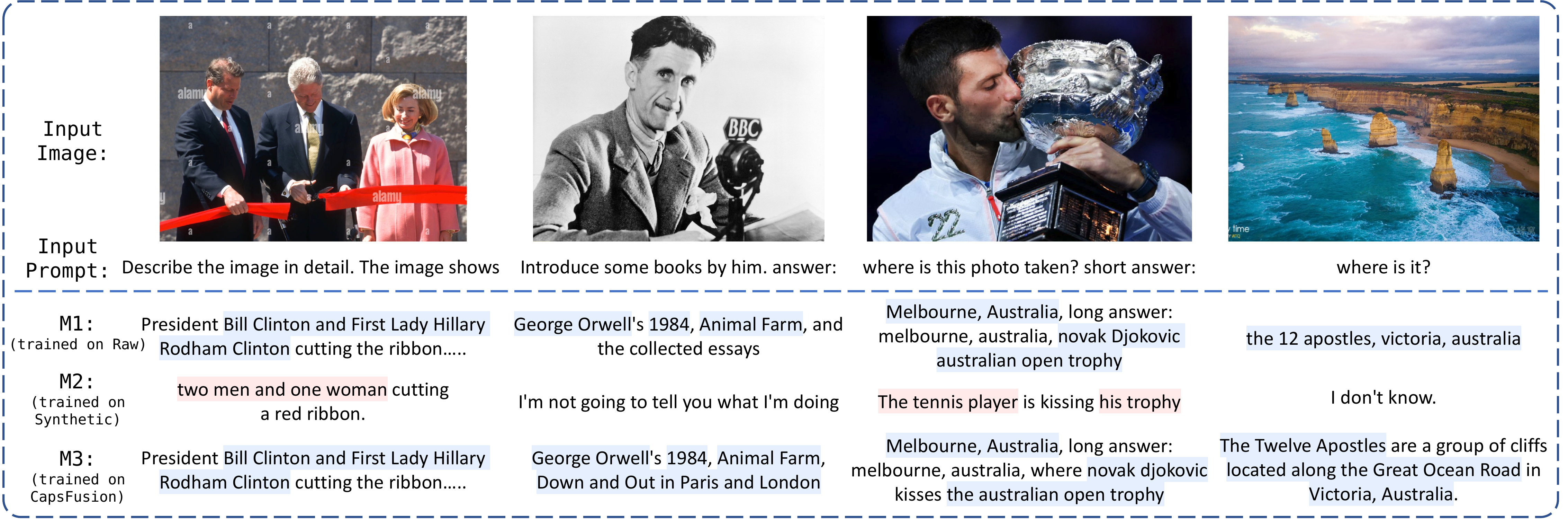}
  \caption{
  Outputs of models trained with different caption datasets. Models trained on raw and \Ours captions (M1 and 3) possess strong world knowledge (in \colorbox{capsblue}{blue}), while the model trained on synthetic captions (M2) can only generate generic concepts (in \colorbox{capsred}{red}).
  }
  \label{fig:intro_model_output_cases}
\end{figure*}

Although achieving promising performance on classic benchmarks such as COCO Caption~\cite{chen2015cococaption}, our further evaluations on recent benchmarks such as SEED-Bench~\cite{li2023seedbench} reveal that training LMMs with large-scale synthetic captions alone is problematic.
We conduct a closer examination of the large-scale training process of LMMs and observe that model training on synthetic captions rapidly reaches a saturation point, beyond which the model performance may even degrade (as illustrated by the green lines in \cref{fig:intro_scale_def}). 
While this severe \textit{Scalability Deficiency} may not be readily apparent on traditional benchmarks such as COCO caption (\cref{fig:intro_scale_def}-a), it becomes notably pronounced  (\cref{fig:intro_scale_def}-b) on the new benchmark SEED-Bench, which supports a  much more comprehensive assessment of LMMs than COCO.
We conduct further analysis on the generated outputs from different models trained with captions of varying quality. \cref{fig:intro_model_output_cases} illustrates system responses trained on Raw captions (M1), Synthetic captions (M2), and our captions (M3).
These examples demonstrate that the outputs from M2, in particular, suffer from severe \textit{World Knowledge Loss}, constituting only high-level concepts while missing all the details about well-known people, locations, events, etc.
The generated sentences by M3 (trained on our captions) are more natural and semantically richer than those from M1 and M2. 

Through examining the differences between raw caption data and synthetic data used in training, we observe that the simplistic syntactic and semantic structures in synthetic captions (\cref{fig:intro_knowledge}-a) may have potentially attributed to the \textit{Scalability Deficiency} and \textit{World Knowledge Loss} issues, which so far have been obscured by their initial benchmark success.
The root cause is that currently used captioning models (\eg BLIP~\cite{li2022blip} used in LAION-COCO~\cite{laioncoco}) for generating synthetic captions heavily rely on academic datasets such as COCO and Conceptual Captions~\cite{sharma2018cc3m} for training. 
These datasets replace specific details (\eg people's names, locations,  landmarks) with more generic \textit{conceptual} placeholders (\eg `person', `city') in the data collection process (\cref{fig:intro_knowledge}-b). Although this eases the training of captioning models, it inevitably results in the loss of a substantial reservoir of valuable real-world information in the trained model, which learns an overly-simplified language structure with basic semantics.
Consequently, LMMs trained on the synthetically simplified datasets generated by these captioning models suffer from a deficiency in language complexity and knowledge depth. 

\begin{figure*}[t]
  \centering
    \includegraphics[width=\linewidth]{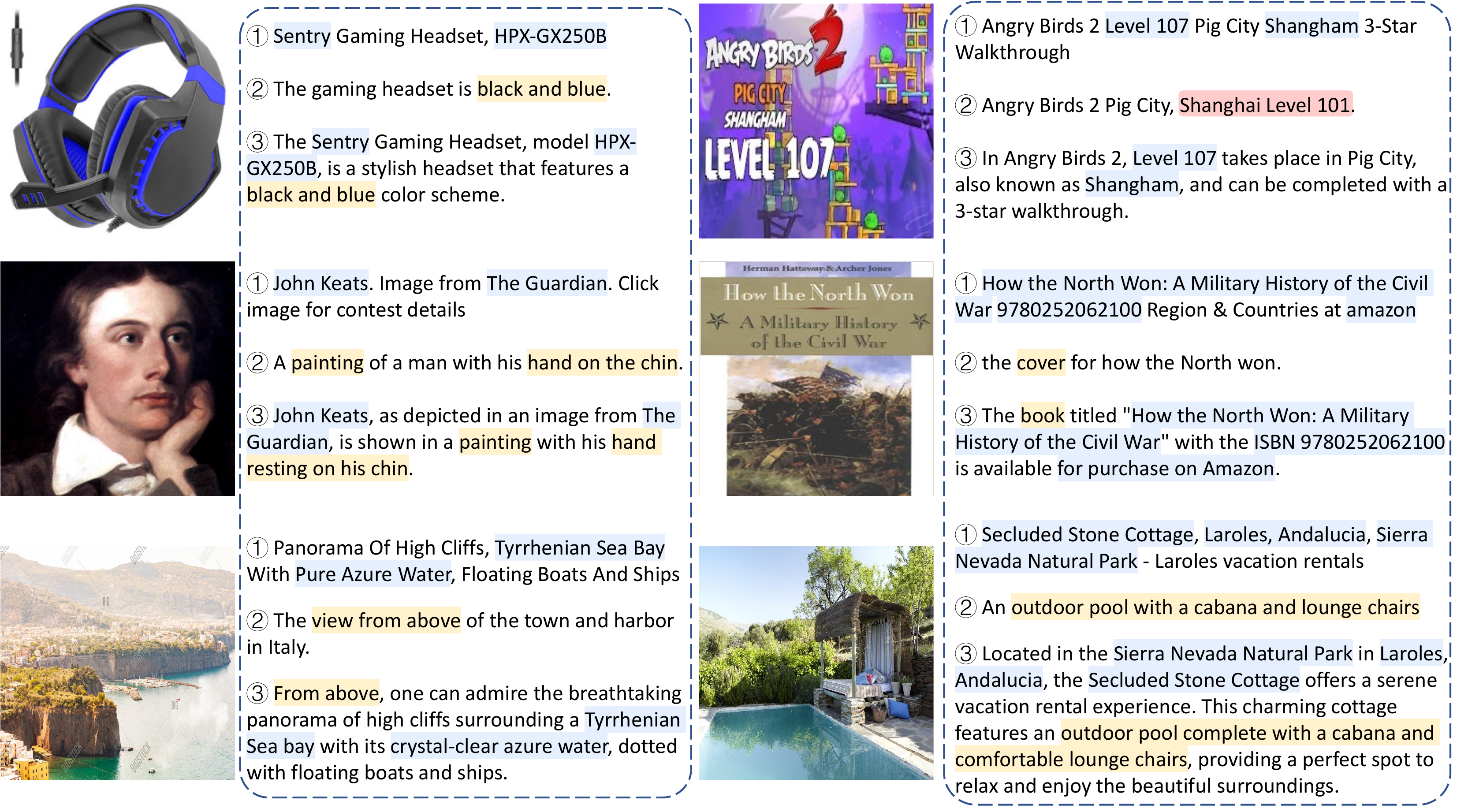}
  \caption{Examples of \ding{192} raw captions (from LAION-2B), \ding{193} synthetic captions (from LAION-COCO, generated by BLIP), and their corresponding \ding{194} \Ours captions. Knowledge from raw captions (in \colorbox{capsblue}{blue}) and information from synthetic captions (in \colorbox{capsyellow}{yellow}) are organically fused into integral \Ours captions. \Ours captions can also correct false information in synthetic captions (in \colorbox{capsred}{red}). More examples can be found in \cref{fig:app_training_caption_cases}.}
  \label{fig:intro_training_caption_cases}
\end{figure*}

Therefore, to train a scalable LMM with abundant real-world knowledge, it is crucial to develop an effective strategy to better synthesize caption data while distilling real-world knowledge from web-based image-text pairs.
There have been some recent attempts to leverage both raw and synthetic captions straightforwardly, by simply mixing them with a fixed hand-tuned ratio~\cite{fan2023laclip,gadre2023datacomp,nguyen2023improvingmmcaptioning}. 
In this work, we propose \Ours, a more advanced pipeline that leverages large language models (LLMs) to enhance the quality of large-scale image-text data. 
\Ours first uses a captioning model~\cite{li2022blip} (following~\cite{laioncoco,li2022blip}) to generate synthetic captions for images. 
Then, 
it utilizes ChatGPT~\cite{schulman2022chatgpt} to 
organically integrate raw and synthetic captions, by extracting real-world knowledge from the structure-flawed raw captions while merging with  structured but syntactically simplified synthetic captions. 
Our evaluations show that ChatGPT excels in this task, but is non-scalable due to its restrictive access. 
To overcome this limitation, we use the outputs generated by ChatGPT as training data to finetune a LLaMA~\cite{touvron2023llama2}.
Evaluation of this finetuned, task-specific LLM demonstrates that it performs \textit{on par with} ChatGPT and consistently produces high-quality captions, while easy to scale up. The trained model is then employed for large-scale caption fusion (examples are presented in \cref{fig:intro_training_caption_cases}, which clearly demonstrate the advantages of \Ours). 

Extensive experiments show that \Ours captions demonstrate remarkable all-around superiority, as a better substitute for both synthetic and raw captions in the training of LMMs.
In terms of \textit{model performance}, \Ours captions clearly outperform synthetic captions by substantial margins, with an improvement of \evagreen{18.8}, \evagreen{18.3}, \evagreen{19.7}, and \evagreen{15.6} in CIDEr score on COCO, NoCaps, TextCaps, and Flickr30K datasets, respectively.  This compelling advantage extends to \textit{sample efficiency} as well. Refined captions from \Ours require \evagreen{11-16} times less computation to achieve  high performance similar to synthetic captions. Furthermore, our investigation unveils that \Ours captions  surpass raw captions by a considerable margin when evaluated on \textit{world knowledge}. Also importantly, \Ours captions demonstrate greater \textit{scalability}, meaning that model performance continues to improve with an increased volume of training samples. This scalability advantage, critical for the training of large-scale models, positions \Ours as a promising candidate for further scaling efforts in LMM training.

\section{Related Work}

\begin{figure*}[t!]
  \centering
    \includegraphics[width=0.9\linewidth]{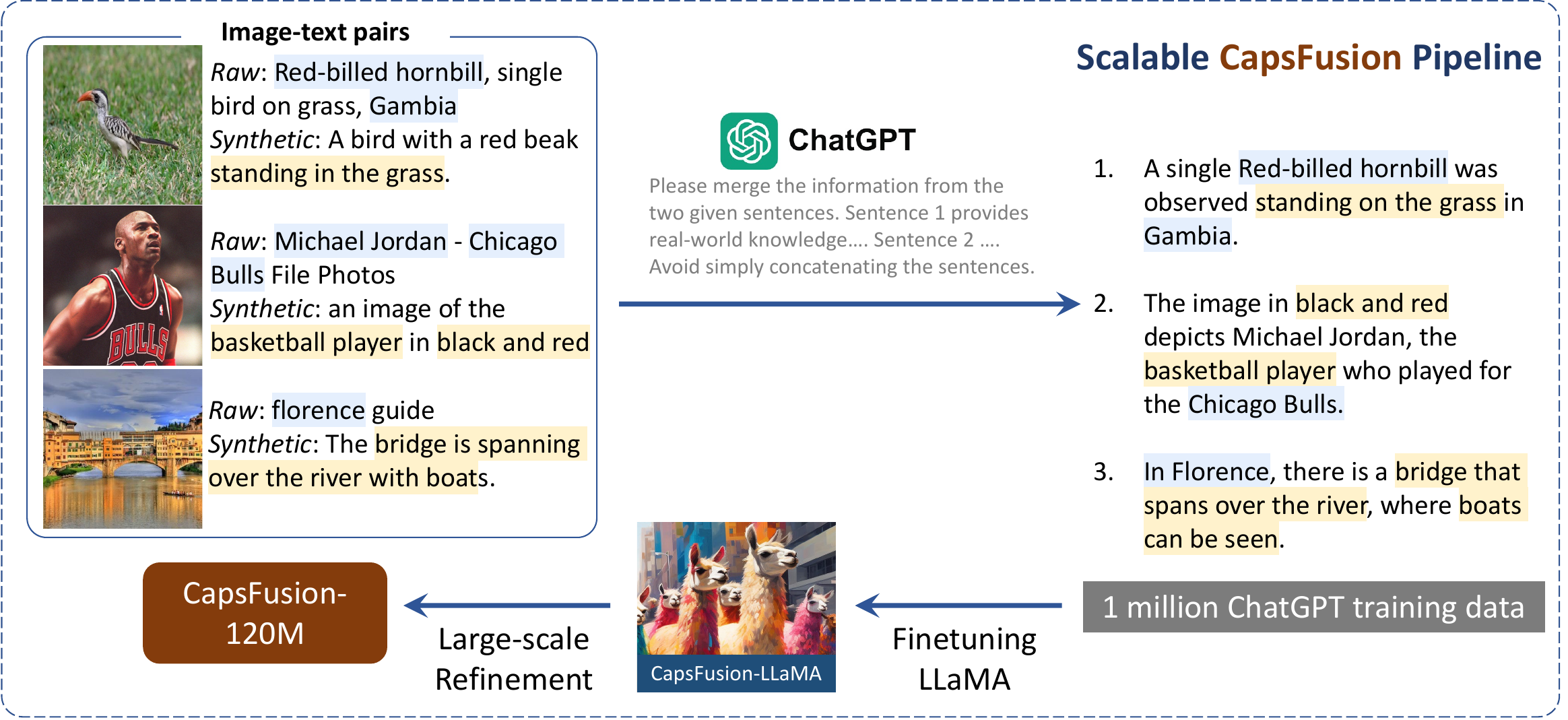}
  \caption{Illustration of the scalable \Ours pipeline for generating high-quality large-scale image-text data.}
  \label{fig:method}
\end{figure*}

\paragraph{Image-text Data Enhancement} 
LaCLIP~\cite{fan2023laclip} utilizes LLM to rewrite raw captions, whose performance can be limited due to severe hallucination, because of limited visual information and low-quality raw captions.
\cite{gadre2023datacomp,nguyen2023improvingmmcaptioning} investigate how to filter and then mix raw and synthetic captions to induce a better CLIP model~\cite{radford2021clip}. 
FuseCap~\cite{rotstein2024fusecap} uses visual experts like object detector to improve captions, while we use large-scale data from the web to enhance.
Our concurrent work VeCLIP~\cite{lai2023veclip} proposes to use LLM to combine information from raw and synthetic captions. The difference is that they directly use an existing LLM for inference, while we finetune a state-of-the-art open-source LLM with training data generated by ChatGPT.
In addition, they have no explicit instructions such as extracting world knowledge present in raw captions and referring sentence structure of synthetic captions, which we use to help LLMs make informed decisions during the caption fusion process.

All recent studies focus on training CLIP models. We instead investigate LMMs and derive insights from a new perspective, such as mixing raw and synthetic captions~\cite{fan2023laclip,nguyen2023improvingmmcaptioning,lai2023veclip} induces no improvement than separate captions. 

\paragraph{Large Multimodal Models} With the success of large language models~\cite{brown2020gpt3,touvron2023llama2} (LLMs), recent studies explore building large multimodal models~\cite{wang2023allseeing,gong2023multimodalgpt,li2023otter,ye2023mplugowl,zhao2023bubogpt,gao2023llamaadapterv2,chen2022pali,zhang2023gpt4roi,chen2023shikra,you2023ferret,chen2023pvitchenchi,peng2023kosmos2,zeng2023lynx,chen2023politeflamingo,huang2023sparkles,zhao2023svit,yao2023deepspeedvisualchat} (LMMs) on LLMs with pretrained vision encoders~\cite{radford2021clip,sun2023evaclip,wang2023whatmakes}.
Most existing works commonly use the prediction of the next text token as the objective~\cite{li2023blip2,instructblip,huang2023kosmos-1,liu2023llava}. Another type of LMMs learns to predict both image and text tokens~\cite{sun2023emu,ge2023seed,yu2023cm3leon,dong2023dreamllm}, endowing models with more versatile abilities of processing both text and image generation tasks, while maintaining image-to-text performance comparable to LMMs trained with only text token supervision.

\section{\Ours}

Large Multimodal Models~\cite{alayrac2022flamingo,li2023blip2,jin2023lavit,wang2023switchgpt,su2023pandagpt,zhang2023vpgtrans,zhu2023minigpt4} serve as a powerful generalist for diverse multimodal tasks. Typical LMM generalist unifies image-to-text tasks only (\eg image captioning and visual question answering). Recent studies such as Emu~\cite{sun2023emu} further enhance the capabilities of multimodal generalist by enabling it to perform both image-to-text and text-to-image tasks in a zero-shot manner~\cite{koh2023FROMAGe,koh2023GILL,ge2023seed-llama,zheng2023minigpt5,wu2023nextgpt}.

\noindent\myparagraph{Learning Objective of LMM.}
The LMM generalist ability originates from a GPT-style auto-regressive training objective~\cite{radford2018gpt1}, wherein the model learns to predict the next token in a sequence. 
As a result of this training paradigm, during inference, the model exhibits a remarkable capability to generate appropriate completions for a wide range of tasks.

Image-text pairs are the most commonly used multimodal pretraining data for learning vision-language alignment.
Specifically, given a dataset $\mathcal{D}$ consisting of image-text pairs $(I, T)$, where $I$ represents the image and $T$ represents the text represented by a sequence of text tokens $T = \{t_1, t_2, \dots, t_n\}$.
The typical training objective is maximizing the conditional likelihood of text tokens $T$ given $I$ in an auto-regressive manner:
\begin{align}
    \max_\theta \frac{1}{|\mathcal{D}|} \sum_{(I, T)\in\mathcal{D}} \sum_{i=1}^{n} 
    \log
    P(t_i|t_1, \dots, t_{i-1}, I; \theta)
\end{align}
Under this training objective, the presence of noisy captions can lead the model to generate extraneous words. Conversely, if the captions are overly simplistic in nature, the model may learn a simplified output style, resulting in a loss of language complexity. Therefore, high-quality image-text pairs are in urgent need  to power new-generation LMMs.

\noindent\myparagraph{Caption Generation.}
Given raw image-text pairs, \Ours first generates synthetic captions using image captioning models following~\cite{li2022blip,laioncoco}. 
In previsou analysis (\cref{fig:intro_scale_def,fig:intro_model_output_cases,fig:intro_knowledge}), we find that raw captions contain a wealth of real-world knowledge but are noisy, while synthetic captions have clean structures but lack in-depth real-world knowledge, which exhibits severe scalability issues.
Thus, our objective is to develop a scalable framework to organically integrate information from both raw and synthetic captions, to create a comprehensive refined image-text dataset.

\begin{figure*}[t]
  \centering
    \includegraphics[width=0.9\linewidth]{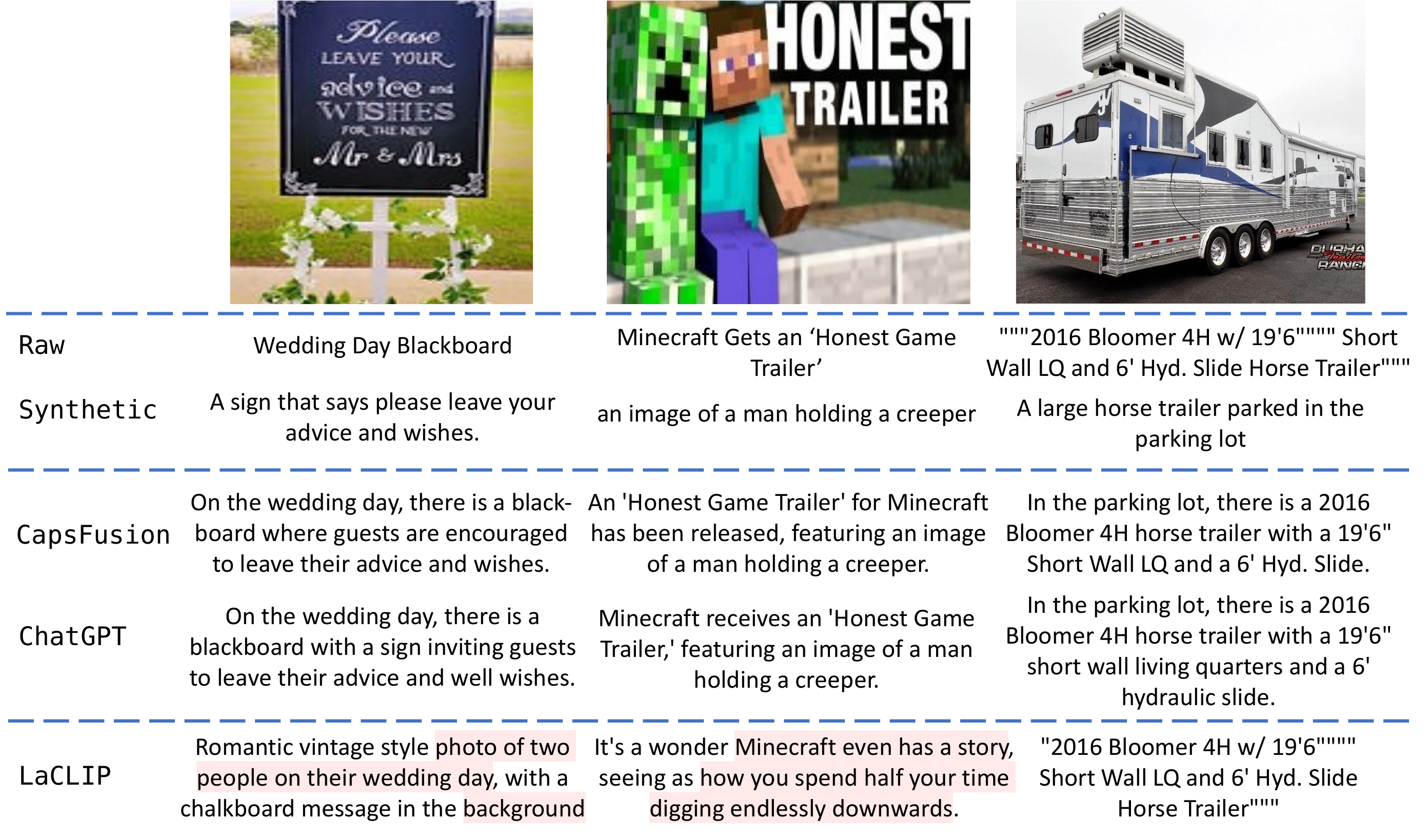}
  \caption{Comparison among \Ours-LLaMA, ChatGPT, and LaCLIP. \Ours-LLaMA performs on par with ChatGPT on the caption fusion task, while LaCLIP suffers severe hallucination because only raw text is considered (hallucinations are highlighted in \colorbox{capsred}{red} in image 1 and 2). LaCLIP also fails when the raw caption is too noisy, while \Ours-LLaMA and ChatGPT can extract useful information from noise (image 3).}
  \label{fig:comparison_la}
\end{figure*}

\noindent\myparagraph{Caption Fusion via ChatGPT.} 
In \Ours, we use ChatGPT to fuse raw and synthetic captions given a prompt.
The task instruction is structured in three key elements: the task description, caption property, and the desired output specifications.
Specifically, we first include a task description that conveys the following objective to ChatGPT: \emph{Please merge the information from two provided sentences.} Furthermore, we provide the distinct properties of the two captions involved, with the following contextual guidance:
\begin{center}
\emph{Raw captions offer detailed real-world information, yet it suffers from flaws in sentence structure and grammar. Synthetic captions exhibit impeccable sentence structure but often lack in-depth real-world details and may contain false information.}
\end{center}
This nuanced description helps ChatGPT make informed decisions during fusion.
Finally, we outline our expectations for the output captions with the following directive: 
\begin{center}
\emph{Ensure a well-structured sentence while retaining the detailed real-world information provided in the raw caption.}
\end{center}
This guideline succinctly encapsulates the desired characteristics of the generated captions.

In our experimentation, we observe that in a portion of samples, ChatGPT resorts to a straightforward concatenation of the raw and synthetic captions for fusion. To address this, we  explicitly instruct ChatGPT to \emph{avoid simply concatenating two sentences}, a directive we have found highly effective in mitigating this issue. The full instruction template is presented in \cref{sec:prompt}.

During human evaluation, ChatGPT is shown to be exceptionally effective at this caption fusion task. Examples are provided in the fourth row of \cref{fig:comparison_la}. We acquired 1 million fused captions using the \texttt{gpt-3.5-turbo} API.

\begin{table*}[t]
  \centering
  \resizebox{0.98\linewidth}{!}{
  \begin{tabular}{clcccccccccc}
  \toprule
  \multirow{2}{*}{Scale} & \multirow{2}{*}{Captions} & \multicolumn{2}{c}{COCO} & \multicolumn{2}{c}{NoCaps} & \multicolumn{2}{c}{TextCaps} & \multicolumn{2}{c}{Flickr30K} \\\cmidrule(lr){3-4} \cmidrule(lr){5-6} \cmidrule(lr){7-8}\cmidrule(lr){9-10} \cmidrule(lr){11-11}
  & & SPICE & CIDEr & SPICE & CIDEr & SPICE & CIDEr & SPICE & CIDEr \\\midrule[0.5pt]
    \multirow{4}{*}{10M} 
  & Raw~\cite{schuhmann2022laion5b} & 15.5 & 75.1 & 9.0 & 64.0 & 10.5 & 46.4 & 13.6 & 54.4 \\
  & Synthetic~\cite{laioncoco} & 19.8 & 102.5 & 11.7 & 84.2 & 12.7 & 42.3 & 15.0 & 63.9 \\
  & Language Rewrites~\cite{fan2023laclip} & 14.6 & 71.6 & 8.6 & 59.0 & 9.3 & 38.3 & 11.6 & 49.0 \\
  & Mixing (Raw \& Syn.)~\cite{fan2023laclip,nguyen2023improvingmmcaptioning} & 17.9 & 90.5 & 10.6 & 76.7 & 12.4 & 51.7 & 15.1 & 64.0  \\
  & Mixing (Raw \& LR)~\cite{fan2023laclip} & 15.0 & 72.6 & 9.0 & 61.1 & 10.3 & 44.6 & 12.2 & 51.7 \\
  & \Ours & \textbf{20.7}\dtplus{+0.9} & \textbf{107.7}\dtplus{+5.2} & \textbf{12.6}\dtplus{+0.9} & \textbf{92.4}\dtplus{+8.2} & \textbf{13.9}\dtplus{+1.2} & \textbf{56.3}\dtplus{+4.6} & \textbf{15.9}\dtplus{+0.8} & \textbf{68.4}\dtplus{+4.4} \\ \midrule[0.5pt]
\multirow{4}{*}{50M} 
  & Raw~\cite{schuhmann2022laion5b} & 16.4 & 81.0 & 9.7 & 68.4 & 11.7 & 55.2 & 14.3 & 60.3 \\
  & Synthetic~\cite{laioncoco} & 19.2 & 100.9 & 11.5 & 82.5 & 13.2 & 46.7 & 14.3 & 60.2 \\
  & Mixing (Raw \& Syn.)~\cite{fan2023laclip,nguyen2023improvingmmcaptioning} & 18.5 & 93.3 & 10.9 & 79.7 & 12.7 & 55.5 & 15.1 & 64.6 \\
  & \Ours & \textbf{21.3}\dtplus{+2.1} & \textbf{112.4}\dtplus{+11.5} & \textbf{13.6}\dtplus{+2.1} & \textbf{99.2}\dtplus{+16.7} & \textbf{14.9}\dtplus{+1.7} & \textbf{62.7}\dtplus{+7.2} & \textbf{16.9}\dtplus{+1.8} & \textbf{74.5}\dtplus{+9.9} \\\midrule[0.5pt]
  \multirow{4}{*}{100M} & Raw~\cite{schuhmann2022laion5b} & 17.1 & 85.5 & 10.1 & 72.8 & 12.3 & 59.6 & 14.6 & 62.2 \\
  & Synthetic~\cite{laioncoco} & 18.5 & 96.9 & 11.0 & 81.6 & 13.1 & 46.5 & 13.7 & 57.4 \\
  & Mixing (Raw \& Syn.)~\cite{fan2023laclip,nguyen2023improvingmmcaptioning} & 18.0 & 95.0 & 10.5 & 77.9 & 12.3 & 55.1 & 15.0 & 66.5  \\
  & \Ours & \textbf{21.7}\dtplus{+3.2} & \textbf{115.7}\dtplus{+18.8} & \textbf{13.5}\dtplus{+2.5} & \textbf{99.9}\dtplus{+18.3} & \textbf{15.2}\dtplus{+2.1} & \textbf{66.2}\dtplus{+11.0} & \textbf{16.8}\dtplus{+1.8} & \textbf{73.0}\dtplus{+6.4} \\
  \bottomrule
  \end{tabular}}
  \caption{Zero-shot evaluation of models trained with different caption datasets on a broad range of image captioning benchmarks.}
  \vspace{-1em}
  \label{tab:result_caption}
\end{table*}

\noindent\myparagraph{Refinement Model with Fused Caption.}
Although ChatGPT is effective, time and computational costs are prohibitive. For scaling, 
we opt to employ LLaMA-2~\cite{touvron2023llama2}, a state-of-the-art open-source LLM.
We finetune the 13B version of LLaMA-2 specifically for the task of caption fusion, using triplets obtained from ChatGPT. These triplets consist of raw and synthetic captions as inputs, with \Ours captions as the target outputs. Training hyperparameters can be found in \cref{sec:train_hyper}. The finetuned model, referred to as \model, is rigorously evaluated through human evaluation on 100 validation cases. The evaluation results are presented in \cref{tab:human_eval}, revealing that the performance of the finetuned \model performs on par with ChatGPT, with 80 out of 100 samples performing equally or better. 
LaCLIP~\cite{fan2023laclip} also leverages LLM for enhancing image-text captions, but simply asks LLM to rewrite raw captions. 
Qualitative comparisons among LaCLIP, \Ours, and ChatGPT are illustrated in Figure \ref{fig:comparison_la}. Notably, LaCLIP tends to hallucinate information not present in the associated image, due to the absence of detailed visual information represented in the raw captions. 
On the other hand, \model exhibits outputs similar to ChatGPT and delivers exceptional performance.
\begin{table}[h]
  \centering
  \resizebox{\linewidth}{!}{
  \begin{tabular}{@{}ccccc@{}}
    \toprule
    & \makecell{ChatGPT\\win} & \makecell{\model\\win} & \makecell{Similar\\quality} & \makecell{(Nearly)\\Identical} \\
    \midrule
    Number & 20 & 15 & 46 & 19 \\
    \bottomrule
  \end{tabular}
  }
  \caption{Human evaluation on \model vs.\ ChatGPT over 100 validation samples.} 
  \vspace{-1em}
  \label{tab:human_eval}
\end{table}

\noindent\myparagraph{Large-scale Caption Fusion.}
The trained \model, being as effective as ChatGPT, now possesses the ability to organically fuse and harness raw and synthetic captions in a manner that is both scalable and highly effective.
We randomly select a subset containing 127,897,754 image-text pairs from LAION-COCO~\cite{laioncoco}, which contains both raw captions from the web and synthetic captions generated by BLIP~\cite{li2022blip}. Subsequently, we apply \model to organically integrate the captions of these image-text pairs. This process costs about 12 days using 128 A100-40G GPUs. After filtering with heuristic rules, we retain a total of 120,724,312 image-text pairs, which we term as the \dset dataset.

\noindent\myparagraph{\dset Dataset.}
\cref{tab:stat} provides a comparison of \dset with existing image-text datasets. We compute the number of unique trigrams and the average length of these captions (word instead of token as unit) in each dataset.
Notably, \dset exhibits the highest count of unique trigrams and the longest average sentence length, underscoring superb diversity within its captions. In contrast, synthetic captions (LAION-COCO) exhibit a considerably lower number of trigrams, signifying a notable lack of language complexity. 

\begin{table}[h]
  \centering
  \begin{tabular}{@{}lcc@{}}
    \toprule
    Datasets & \# Unique Trigrams & Avg. Length \\
    \midrule
    LAION-2B & 5.51 $\times 10^7$ & 10.95 \\
    LAION-COCO & 1.00 $\times 10^7$  & 8.99 \\
    La-CLIP & 5.46 $\times 10^7$ & 14.63\\
    \dset & 7.13 $\times 10^7$ & 22.74 \\\bottomrule
  \end{tabular}
  \caption{Statistics of different caption datasets (on a randomly selected 10 million subset of \dset images).}
  \label{tab:stat}
\end{table}

\begin{figure*}[t]
  \centering
\includegraphics[width=1.0\linewidth]{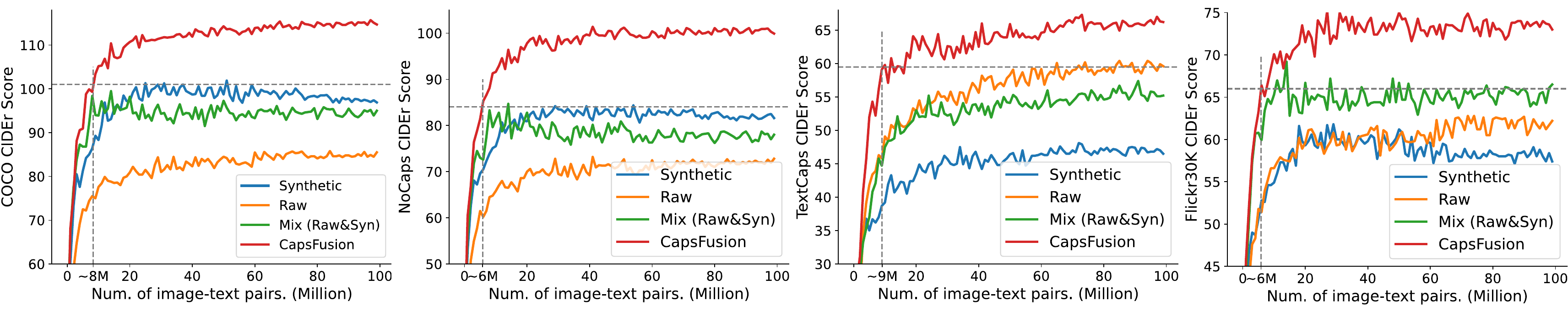}
  \caption{Comparison of scalability and sample efficiency  across different datasets.}
  \vspace{-1em}
  \label{fig:exp_training_process}
\end{figure*}

\section{Experiments}

We present a comprehensive analysis of different caption datasets. Extensive experiments show that \Ours exhibits all-around superiority over existing image-text pair datasets, in terms of effectiveness, efficiency, world knowledge depth, and scalability.

\subsection{Setup}

For a fair comparison, we compare \Ours with other caption datasets under the same set of images from LAION-COCO~\cite{laioncoco}, isolating caption quality as the only varying factor.
Experiments are conducted across three scales: 10, 50 and 100 million image-text pairs.

\noindent\myparagraph{Model Architecture.} 
We adopt the most prevalent LMM architecture, consisting of three components: an LLM, a vision encoder, and a vision-language bridging module.
We use LLaMA-2-7B~\cite{touvron2023llama2} and EVA-01-CLIP-g~\cite{fang2023eva,sun2023evaclip} to initialize the LLM and vision encoder modules, respectively.
For the bridging module, we follow Emu~\cite{sun2023emu} to use a randomly initialized Causal Transformer to bridge the vision and language modalities. This module transforms bi-directional image representations from the vision encoder into a causal sequence that aligns better to the nature of LLMs, which excel at modeling causal sequences in an autoregressive fashion. 
The LLM and vision encoder are frozen during training to save computation cost following~\cite{li2023blip2}, and only the bridging module is tuned.

\noindent\myparagraph{Training Schedule.} 
The training schedule is set as the same for all compared captions.
For each evaluation scale, we train the model for 1 epoch. This practice follows Datacomp~\cite{gadre2023datacomp}, a benchmark for evaluating image-text pair datasets on CLIP training. 
The peak learning rate is 3e-4, with the initial 2,000 (100M) / 1,000 (50M) / 500 (10M) steps as warm-up, after which the learning rate decreases to 3e-5 with a cosine learning rate decay schedule. Batch size is set to 8192 for all scales. Detailed training hyperparameters can be found in \cref{sec:train_hyper}. The 100M scale training costs 40 hours with 16 A800-80G GPUs.

\noindent\myparagraph{Baselines.} 
We establish two baselines using raw captions from LAION-2B~\cite{schuhmann2022laion5b} and synthetic captions from LAION-COCO~\cite{laioncoco}. Additionally, two state-of-the-art methods for improving image-text pairs in CLIP training are evaluated: language rewrites (LaCLIP~\cite{fan2023laclip}) and random mixing~\cite{nguyen2023improvingmmcaptioning,fan2023laclip}.
For \cite{fan2023laclip}, we adopt their in-context strategy and employ LLaMA-2-7B to rewrite 10M captions for comparison, taking 30 hours with 8 A100-40G GPUs.
For random mixing~\cite{fan2023laclip,nguyen2023improvingmmcaptioning}, we set the mixing ratio of two types of captions as 1:1~\cite{fan2023laclip} and do not tune this ratio as in~\cite{nguyen2023improvingmmcaptioning}.

\noindent\myparagraph{Evaluation.}
We comprehensively assess the performance of LMMs across a wide range of evaluation benchmarks. These benchmarks encompass both traditional benchmarks and recently introduced assessments, including COCO~\cite{chen2015cococaption}, NoCaps~\cite{agrawal2019nocaps}, TextCaps~\cite{sidorov2020textcaps}, Flickr30k~\cite{plummer2015flickr30k}, and SEED-Bench~\cite{li2023seedbench}. 
For image captioning tasks, we employ SPICE~\cite{anderson2016spice} and CIDEr~\cite{vedantam2015cider} metrics.
For the comprehensive SEED-Bench in the form of multiple-choice questions, we evaluate LMMs using accuracy.

\subsection{Model Performance}
The performances of models trained with different captions on COCO, NoCaps, TextCaps, and Flickr30K benchmarks are presented in \cref{tab:result_caption}. 
We observe that \Ours outperforms all baseline captions in all settings by a large margin, across all datasets evaluated. 
For example, on the 100M scale, \Ours surpasses the best baseline by a substantial margin, achieving 18.8 and 18.3 CIDEr score improvements on COCO and NoCaps, respectively.

\begin{table*}[t]
  \centering
  \resizebox{\linewidth}{!}{
  \begin{tabular}{lcccccccccc}
  \toprule
  Captions & Scene U & Inst Iden & Inst Loc & Inst Attr & Inst Cnt & Spatial Rel & Inst Inter & Vis Reason & Text Rec & Total \\\midrule[0.5pt]
  Raw~\cite{schuhmann2022laion5b} & 57.9 & 51.2 & 39.8 & 47.7 & 44.6 & 35.3 & 47.4 & \textbf{48.6} & \textbf{34.1} & 48.7 \\
  Synthetic~\cite{laioncoco} & 52.7 & 48.9 & 36.7 & 42.2 & 35.7 & 34.5 & 48.4 & 35.0 & 12.9 & 43.2 \\
  \Ours & \textbf{58.8} & \textbf{52.7} & \textbf{41.0} & \textbf{48.0} & \textbf{46.3} & \textbf{35.9} & \textbf{57.7} & 47.1 & 20.0 & \textbf{49.8} \\
  \bottomrule
  \end{tabular}
  }
  \caption{Zero-shot evaluation of models trained with different caption datasets on SEED-Bench.}
  \vspace{-1em}
  \label{tab:seed-bench}
\end{table*}

\noindent\myparagraph{Rewriting Captions Fails at Image Captioning.} 
On the 10M scale, our examination reveals that Language Rewrites captions~\cite{fan2023laclip}, generated through the process of rewriting raw captions, fail to achieve decent performance. This can be attributed to the severe hallucination issue we observed in the rewrites captions (\cref{fig:comparison_la}), which introduces extraneous text that is irrelevant to the content depicted in the accompanying images. The underlying cause of the hallucination phenomenon can be traced back to the input data, which consists solely of noisy raw captions, providing a suboptimal starting point for the rewriting process.

\noindent\myparagraph{Mixing Captions does not Bring Consistent Gains.}
Another notable observation is that mixing captions cannot yield better performance. For instance, on the 10M-scale over COCO benchmark, mixing raw and LR captions (72.62 CIDEr and 15.01 SPICE scores) achieves a median performance between Raw (75.13 CIDEr and 15.48 SPICE) and LR (71.61 CIDEr, 14.6 SPICE) captions.  
This finding is contrarian to the observation in CLIP training~\cite{lai2023veclip,fan2023laclip,nguyen2023improvingmmcaptioning}, where mixing raw and generated captions has proven to be a strong strategy for enhancing CLIP performance, with raw captions being an indispensable component~\cite{lai2023veclip,fan2023laclip}. In contrast, our experiments show that in LMMs training, the exclusive use of a single caption type (\Ours) can outperform both raw and synthetic captions.

\noindent\myparagraph{Synthetic Captions Shout at Small Scale.}
A noteworthy observation is that synthetic caption demonstrates exceptional results on the 10M dataset (102.5 COCO CIDEr), while exhibiting inferior performance (96.93 COCO CIDEr) on the larger-scale 100M dataset. This aligns with our earlier observation of the \textit{Scalability Deficiency} issue in synthetic captions, a potential threat to the effective training of LMMs.
But even at small scales, the effectiveness of synthetic captions consistently falls behind that of \Ours across all datasets.

\subsection{Sample Efficiency}

In addition to comparing performance across different dataset scales, we probe deeper into training sample efficiency. In \cref{tab:result_caption}, we find that with only 10M image-text pairs, \Ours captions outperform other captions with much larger scale (50M and 100M), demonstrating exceptional sample efficiency. We visualize the updates of evaluation metrics on NoCaps, TextCaps, Flickr30K, and COCO benchmarks when the number of seen training samples increases from 0 to 100 million image-text pairs, presented in \cref{fig:exp_training_process}.
The horizontal grey dashed lines approximately represent the best-saturated performance of baseline captions when trained with 100 million image-text pairs. The vertical dashed line reveals the number of samples employed by \Ours to achieve a similar level of performance as the best-performing baseline captions.
It is worth noting that \Ours attains the same level of performance as the best baseline captions with only 6M, 9M, 6M, and 8M samples for NoCaps, TextCaps, Flickr30K, and COCO captions, respectively. 
This achievement underscores \Ours's ability of 11-16 times speedup and demonstrates its superior sample efficiency.

\subsection{Scalability Analysis}

Scalability stands as a crucial attribute in large model training. 
Our investigation reveals that synthetic captions, among all the caption types considered, exhibit the worst scalability. This can be observed from \cref{fig:exp_training_process} (a), (b), and (d), wherein the blue lines exhibit early saturation with a mere 30 million image-text pairs. Subsequently, their performance gradually deteriorates. 
In contrast, raw caption (orange lines) displays commendable scalability, with its performance showing a consistent upward trajectory.
However, the inherent high noise level in raw caption hampers its ability to achieve strong performance.
\Ours caption (red lines) exhibits remarkable scalability on all datasets, outperforming both synthetic and raw captions by a substantial margin throughout the entire scale.

\noindent\myparagraph{Note:} Our investigation reveals that synthetic captions have severe scalability limitations and typically saturate with only 30 million pairs, after which \textit{more computation imposes an adverse impact on model performance}.
However, current synthetic caption datasets used are typically much larger in scale (\eg 600M in LAION-COCO). We hope our findings raise concerns about the efficiency issue in training LMMs with such massive synthetic caption datasets.

\subsection{Further Evaluation on SEED-Bench}

Recently, new comprehensive benchmarks are proposed for thorough evaluations of LMMs on granular functionalities~\cite{fu2023mme,liu2023mmbench,yu2023mmvet,bai2023touchstone,bitton2023visitbench,xu2023lvlmehub}. 
We evaluate our proposed model on a representative benchmark, SEED-Bench~\cite{li2023seedbench}, over its 9 image-text tasks (dataset details can be found in \cref{app:seed}.). Results are presented in \cref{tab:seed-bench}. We find \Ours outperforms raw and synthetic captions in 7 out of 9 evaluated tasks, which underscores the remarkable capabilities of \Ours in instance counting, instance interaction, scene understanding and other multimodal functionalities.


\begin{table}
  \centering
  \resizebox{\linewidth}{!}{
  \begin{tabular}{@{}lccc@{}}
    \toprule
    Method & COCO~\cite{chen2015cococaption} & SEED-Bench~\cite{li2023seedbench} & MMLU~\cite{hendrycks2020mmlu} \\
    \midrule
    LLaMA-2-7B & - & - & \textbf{45.78} \\ \midrule[0.01em]
    Raw & 74.9 & 48.5 & 43.7 \\
    Synthetic & 55.5 & 40.2 & 42.9 \\
    \Ours & \textbf{111.3} & \textbf{51.5} & \underline{44.1} \\
    \bottomrule
  \end{tabular}
  }
  \caption{Model performance with LLM tuned over different caption datasets.}
  \vspace{-1em}
  \label{tab:exp_language}
\end{table}

\subsection{Qualitative Evaluation on World Knowledge}

\begin{figure*}[t]
  \centering
    \includegraphics[width=\linewidth]{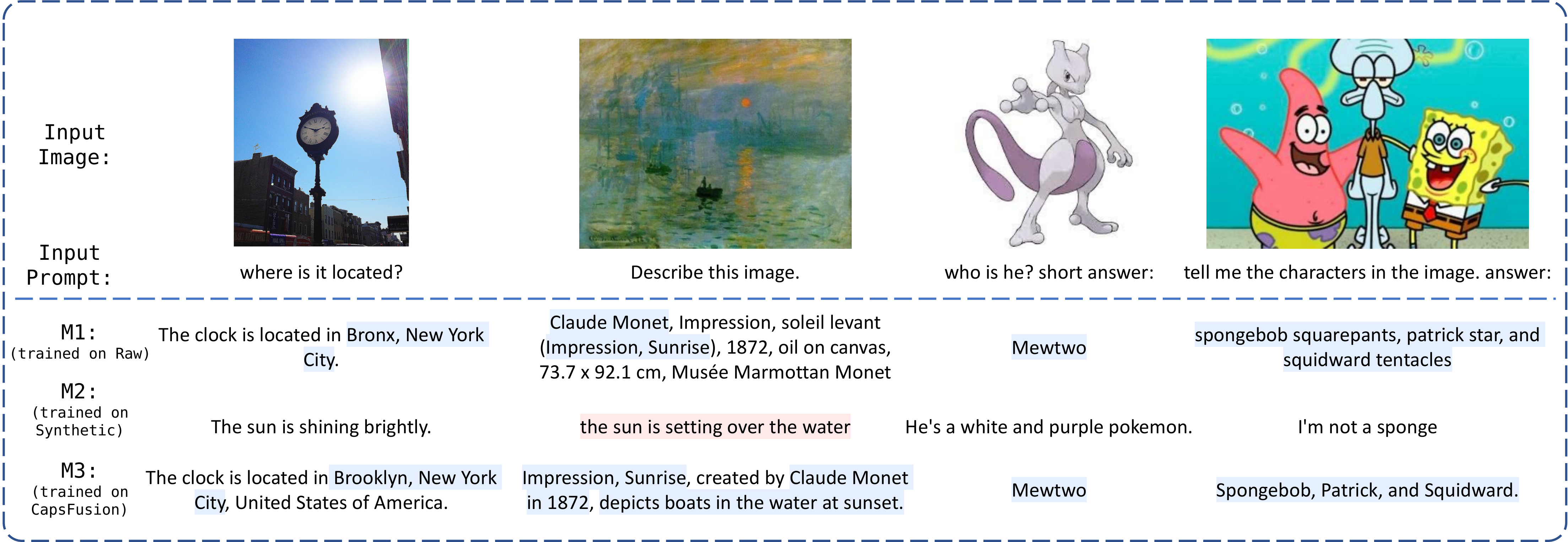}
  \caption{Outputs of models trained with different caption datasets. Models trained on raw and \Ours captions (M1 and 3) possess strong world knowledge (in \colorbox{capsblue}{blue}), while the model trained on synthetic captions (M2) can only generate generic concepts (in \colorbox{capsred}{red}).}
  \label{fig:app_output_cases}
\end{figure*}

In \cref{fig:intro_model_output_cases} and \cref{fig:app_output_cases} (Appendix),  
we provide a qualitative evaluation on the outputs generated by models trained with different datasets. The first row is the input image with text prompt, the lower three rows show the outputs from models trained on raw, synthetic, and \Ours captions.

We observe that models trained on raw and \Ours captions exhibit rich real-world knowledge, able to identify celebrities (\cref{fig:intro_model_output_cases} image 1), recognize famous artworks (\cref{fig:app_output_cases} image 2), attribute literature works to their authors (\cref{fig:intro_model_output_cases} image 2), and pinpoint the location where the specific event occurred (\cref{fig:intro_model_output_cases} image 3). Models trained on synthetic captions totally lost such capabilities.

\subsection{Effects when Firing LLM}

We investigate the impact of different training captions when firing the LLM. We conduct experiments at the 10M scale on COCO~\cite{chen2015cococaption}, SEED~\cite{li2023seedbench}, MMLU~\cite{hendrycks2020mmlu}. Results are summarized in \cref{tab:exp_language}. 
Notably, we observe a significant decline in the performance of synthetic captions. This indicates a deterioration in the LLM's capabilities when it is trained on the simplified language of synthetic captions.

\section{Conclusion}

In this work, we identify severe \textit{Scalability Deficiency} and \textit{World Knowledge Loss} issues in LMMs trained with synthetic captions. On the other hand, web-based image-text pairs possess rich world knowledge but are too noisy to achieve decent performance.
We thus propose \Ours, an advanced framework to generate high-quality captions in a scalable and effective manner. The resulting \dset dataset exhibits all-around superiority over existing image-text datasets,
which poses \Ours as a promising framework to generate large-scale high-quality image-text data for scalable LMM training.

{
    \small
    \bibliographystyle{ieeenat_fullname}
    \bibliography{main}
}

\clearpage
\setcounter{page}{1}
\maketitlesupplementary

\begin{figure*}[t]
  \centering
    \includegraphics[width=\linewidth]{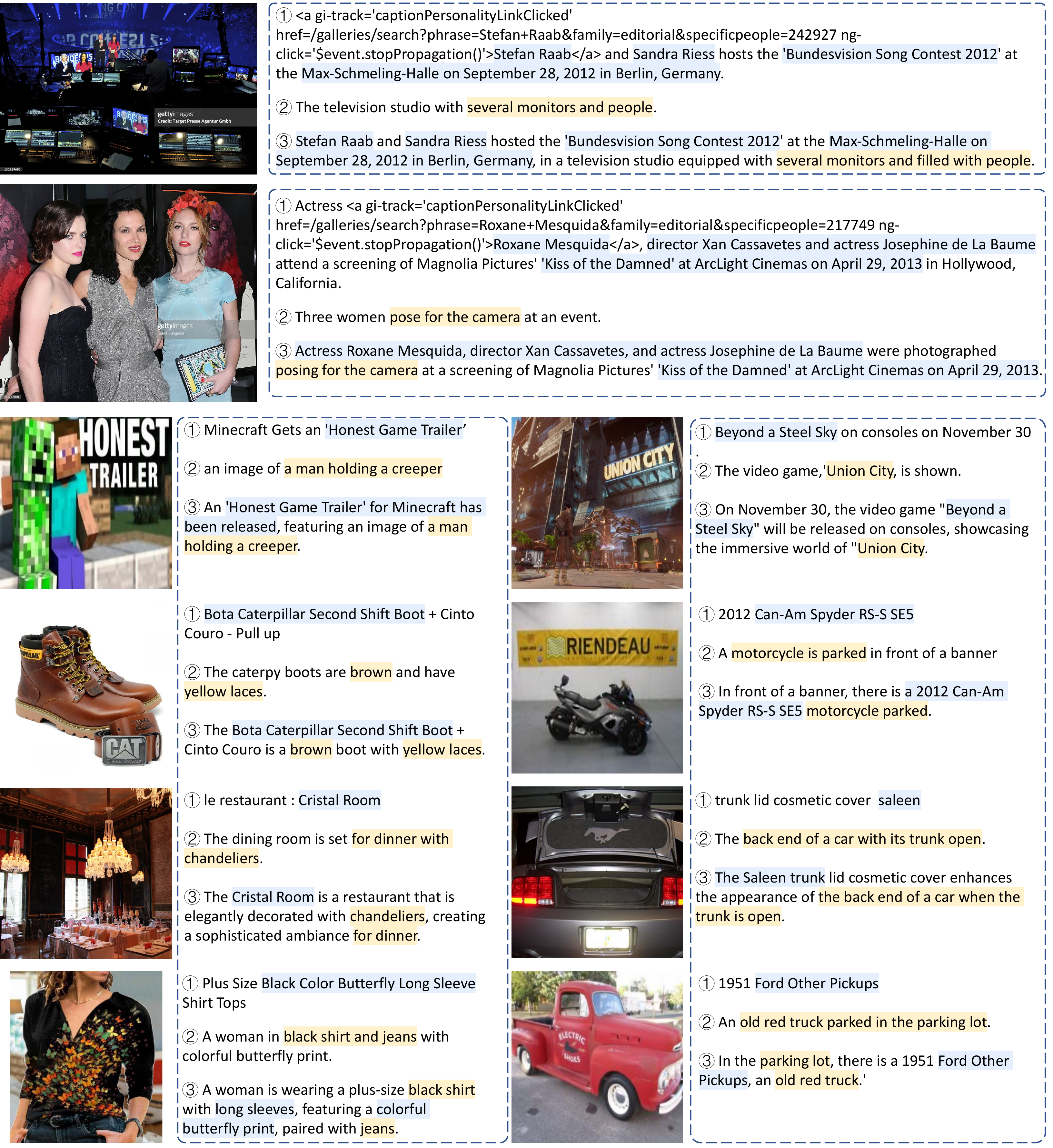}
  \caption{Examples of \ding{192} raw captions (from LAION-2B), \ding{193} synthetic captions (from LAION-COCO, generated by BLIP), and their corresponding \ding{194} \Ours captions. Knowledge from raw captions (in \colorbox{capsblue}{blue}) and information from synthetic captions (in \colorbox{capsyellow}{yellow}) are organically fused into integral \Ours captions. The dataset will be released and more examples can be found there.}
  \label{fig:app_training_caption_cases}
\end{figure*}

\section{More \Ours Examples}

More examples of web-based raw captions, synthetic captions generated by BLIP, and their \Ours captions generated by \model are provided in \cref{fig:app_training_caption_cases}. \Ours captions can organically organize information from raw and synthetic captions.

\section{Prompting Templates for Data Refining}
\label{sec:prompt}

The prompt for ChatGPT and \model to integrate raw and synthetic captions is shown below.

\newcommand{\prompt}[1]{\textcolor{black}{\small{\texttt{#1}}}}
\begin{center}
    %
    \prompt{Please merge and refine the information from the two given sentences.} \\
    \prompt{Sentence 1 provides detailed real-world knowledge, yet it suffers from flaws in sentence structure and grammar.} \\
    \prompt{Sentence 2 exhibits nice sentence structure, but lacking in-depth real-world details and may contain false information.} \\
    \prompt{Please combine them into a new sentence, ensuring a well-structured sentence while retaining the detailed real-world information provided in Sentence 1.} \\
    \prompt{Avoid simply concatenating the sentences.} \\
    \prompt{Sentence 1: <raw caption>} \\
    \prompt{Sentence 2: <synthetic caption>} \\
\end{center}

\section{Hyperparameters}
\label{sec:train_hyper}

Training hyperparameters of \model and LMM are presented in \cref{tab:train_hyper_llama,tab:train_hyper_lmm} respectively.

\begin{table}[t]
    \centering
    \small
    \setlength{\tabcolsep}{0.15cm}
    \renewcommand{\arraystretch}{1.4}
    \begin{tabular}{lc}
    \toprule
    Configuration & \Ours Training \\
    \midrule
    Model init & LLaMA-2-13B \\
    Batch size & 128 \\
    Data & \makecell{1 million (raw, synthetic, fused)\\triplets from ChatGPT} \\
    Training Epoch & 2 \\
    Peak Learning Rate & 1e-5 \\
    End Learning Rate & 0 \\
    Warmup Steps & 500 \\
    LR Scheduler & cosine \\
    Optimizer & AdamW~\citep{loshchilov2017adamw} \\
    Optimizer hyper-parameters & $\beta_1$, $\beta_2$, $\epsilon$ = 0.9, 0.95, 1e-8 \\
    Weight decay & 0.0 \\\bottomrule
    \end{tabular}
    \caption{Summary of \model training hyperparameters.}
    \label{tab:train_hyper_llama}
\end{table}

\begin{table}[t]
    \centering
    \small
    \setlength{\tabcolsep}{0.15cm}
    \renewcommand{\arraystretch}{1.4}
    \begin{tabular}{lc}
    \toprule
    Configuration & Large Multimodal Model Training \\
    \midrule
    Model init & LLaMA-2-13B \\
    Batch size & 8192 \\
    Data & 10 / 50 / 100 million image-text pairs \\
    Training Epoch & 1 \\
    Peak Learning Rate & 3e-4 \\
    End Learning Rate & 3e-5 \\
    Warmup Steps & 2000 \\
    LR Scheduler & cosine \\
    Optimizer & AdamW~\citep{loshchilov2017adamw} \\
    Optimizer hyper-parameters & $\beta_1$, $\beta_2$, $\epsilon$ = 0.9, 0.98, 1e-6 \\
    Weight decay & 0.0 \\\bottomrule
    \end{tabular}
    \caption{Summary of LMM training hyperparameters.}
    \label{tab:train_hyper_lmm}
\end{table}

\section{Details of SEED-Bench}
\label{app:seed}

SEED-Bench~\cite{li2023seedbench} incorporates 12 evaluation tasks including both the spatial and temporal comprehension to comprehensively assess the visual understanding capability of LMMs. We select 9 image-text tasks from them (the left 3 tasks are video-text tasks) for evaluation. The task details are introduced below.

\begin{itemize}
    \item Scene Understanding. This task focuses on the global information in the image. Questions can be answered through a holistic understanding of the image.
    \item Instance Identity. This task involves the identification of a certain instance in the image, including the existence or category of a certain object in the image. It evaluates a model’s object recognition capability.
    \item Instance Location. This task concerns the absolute position of one specified instance. It requires a model to correctly localize the object referred to in the question.
    \item Instance Attributes. This task is related to the attributes of an instance, such as color, shape or material. It assesses a model’s understanding of an object’s visual appearance.
    \item Instances Counting. This task requires the model to count the number of a specific object in the image. This requires the model to understand all objects, and successfully count the referred object’s instances.
    \item Spatial Relation. This task asks an model to ground the two mentioned objects, and recognize their relative spatial relation within the image.
    \item Instance Interaction. This task requires the model to recognize the state relation or interaction relations between two humans or objects.
    \item Visual Reasoning. This task evaluates if a model is able to reason based on the visual information. This requires the model to fully understand the image and utilize its commonsense knowledge to correctly answer the questions.
    \item Text Understanding. For this task, the model should answer question about the textual
elements in the image.
\end{itemize}


\end{document}